\documentclass{article}
\usepackage{url}
\usepackage[pdftex]{graphicx}
\usepackage{amsmath}
\usepackage{booktabs}
\usepackage{amsfonts}
\usepackage{float}
\usepackage{tcolorbox}
\usepackage{natbib}
\usepackage{authblk}
\usepackage{multirow}

\usepackage{authblk}  % Added for better author/affiliation formatting

\title{MechPert: Mechanistic Consensus as an Inductive Bias for Unseen Perturbation Prediction}
\author{Marc Boubnovski Martell\textsuperscript{1}, Josefa Lia Stoisser\textsuperscript{1}, Lawrence Phillips\textsuperscript{1}, Aditya Misra\textsuperscript{1}, Robert Kitchen\textsuperscript{1}, Jesper Ferkinghoff-Borg\textsuperscript{1}, Jialin Yu\textsuperscript{2}, Philip Torr\textsuperscript{2}, Kaspar Märten }

\affil[1]{Novo Nordisk} \affil[2]{University of Oxford}

\date{Feb 2026}

\begin{document}
\maketitle

\begin{abstract} Predicting transcriptional responses to unseen genetic perturbations is essential for understanding gene regulation and prioritizing large-scale perturbation experiments. Existing approaches either rely on static, potentially incomplete knowledge graphs, or prompt language models for functionally similar genes, retrieving associations shaped by symmetric co-occurrence in scientific text rather than directed regulatory logic. We introduce MechPert, a lightweight framework that encourages LLM agents to generate directed regulatory hypotheses rather than relying solely on functional similarity. Multiple agents independently propose candidate regulators with associated confidence scores; these are aggregated through a consensus mechanism that filters spurious associations, producing weighted neighborhoods for downstream prediction. 
We evaluate MechPert on Perturb-seq benchmarks across four human cell lines. For perturbation prediction in low-data regimes ($N=50$ observed perturbations), MechPert improves Pearson correlation by up to 10.5\% over similarity-based baselines. For experimental design, MechPert-selected anchor genes outperform standard network centrality heuristics by up to 46\% in well-characterized cell lines. \end{abstract}

\section{Introduction}

Predicting transcriptomic responses to genetic perturbations remains a fundamental challenge in computational biology, with applications in understanding gene regulation and identifying therapeutic targets \citep{replogle2022mapping, dixit2016perturb}. While high-throughput screens like Perturb-seq have enabled genome-scale profiling, typical experiments remain small to medium-sized, covering only a fraction of the ~20,000 possible single-gene perturbations \cite{shang2025predicting}. The ability to generalize from a small number of observed perturbations to unseen targets (few-shot prediction) has thus emerged as a critical frontier for accelerating biological discovery.

% Predicting transcriptomic responses to genetic perturbations remains a fundamental challenge in computational biology, essential for decoding gene regulatory networks (GRNs) and identifying therapeutic targets \citep{replogle2022mapping, dixit2016perturb}. High-throughput screens like Perturb-seq have generated massive datasets, yet the combinatorial space of possible perturbations—over 20,000 genes across diverse cellular contexts—remains experimentally intractable. The ability to generalize from limited screens to unseen perturbations (\textbf{few-shot prediction}) has thus emerged as the critical frontier for accelerating biological discovery. This represents a critical barrier for the targeted engineering of cellular states.

Current methods attempt to bridge this gap by incorporating prior biological knowledge. Graph Neural Network (GNN) approaches, such as GEARS \citep{roohani2024predicting} and TxPert \citep{wenkel2025txpert}, leverage structured Gene Ontology (GO) graphs to propagate perturbation signals. Language Model approaches like LangPert \citep{martens2025langpert} aggregate contextual information from unstructured literature. While recent Foundation Models such as GRNFormer \citep{qiu2025grnformer} seek to integrate explicit GRN priors into RNA token embeddings, all existing methods share a critical limitation: they rely on static, potentially incomplete knowledge graphs or noisy literature embeddings. When regulatory interactions are missing from curated databases or buried in ambiguous text, these models fail to enforce directional regulatory constraints, leading to poor generalization on understudied genes.

However, whether LLMs are used to generate gene embeddings \citep{chen2024genept} or prompted directly for functional similarity, the resulting associations reflect symmetric co-occurrence in scientific text rather than directed regulatory relationships. If genes A and B are frequently co-mentioned in the literature, both approaches will treat them as similar, without distinguishing whether the directed edge is $A \rightarrow B$, $B \rightarrow A$, or neither\cite{ruan2026large}. This conflation of co-occurrence with regulatory directionality is a distinction central to causal inference \citep{pearl2009causality, martell2025scalable}. As a result, predictions risk mirroring co-citation patterns rather than the directional logic of gene regulatory networks \cite{afonja2024llm4grn}.

% Naive application of LLMs to biological retrieval suffers from a fundamental representation bias: standard language embeddings capture textual co-occurrence ($P(B|A)$) rather than regulatory causality ($P(B|\text{do}(A))$) \citep{pearl2009causality}. This leads to \textbf{structural conflation}, where unconstrained models fail to distinguish upstream drivers (transcription factors) from their downstream targets due to high lexical similarity in scientific corpora \citep{lee2020biobert, beltagy2019scibert}. As a result, predictions often mirror co-citation patterns rather than the true directional mechanisms of gene regulatory networks \citep{chen2024genept}.

To overcome this, we introduce MechPert, a framework that transforms unseen perturbation prediction from similarity-based prompting into a mechanistic reasoning problem. Rather than asking LLMs to identify functionally similar genes as in LangPert \citep{martens2025langpert}, MechPert prompts multiple agents to independently generate directed regulatory hypotheses, which are then aggregated through a consensus mechanism. The framework is algorithmically lightweight, operating entirely at inference time without requiring architectural changes or retraining, and its primary contribution is conceptual: introducing a mechanistic inductive bias that is complementary to model architecture and training scale.

This approach addresses a distinct failure mode from graph-based methods: when curated databases are incomplete but scientific literature is comprehensive, mechanistic reasoning over text provides an alternative to graph-based propagation. While we also explore topological augmentations that benefit smaller language models (e.g., Gemini 3 Flash; Appendix \ref{app:Archit}), we find that for frontier models (e.g., Gemini 3 Pro) the lightweight consensus framework alone is sufficient.

We evaluate \textsc{MechPert} on two complementary tasks using Perturb-seq data across four human cell lines \citep{replogle2022mapping, nadig2024transcriptome}. First, for few-shot perturbation prediction, we demonstrate that directing LLM reasoning toward regulatory relationships improves generalization by a relative \textbf{10.5\%} over similarity-based baselines in low-data regimes ($N=50$). Second, for experimental design, we use \textsc{MechPert} to autonomously identify an initial set of 50 perturbations to map a regulatory landscape. Here, \textsc{MechPert}-selected anchors outperform standard network centrality heuristics by up to \textbf{46\%} in K562, suggesting that LLM-guided target selection can effectively complement structural approaches for experimental prioritization \cite{yan2025comprehensive}.

% We evaluate \textsc{MechPert} on two complementary tasks using the Replogle et al. \citep{replogle2022mapping} benchmark, spanning over 20,000 perturbation pairs across four divergent cell lines. First, on {Few-Shot Perturbation Prediction}, we demonstrate that grounding latent semantics in causal priors improves generalization by a relative \textbf{10.5\%} over baselines in low-data regimes ($N=50$). Second, we address {Active Mechanistic Design} autonomously identifying the optimal initial set of 50 perturbations to map a regulatory landscape. Here, \textsc{MechPert}-selected anchors outperform traditional network centrality heuristics by up to \textbf{46\%} in K562, confirming that agentic reasoning can effectively guide experimental design in combinatorial biological spaces.

%%%%%%%%%%%%%%%%%%%%%%%%%%%%%%%%%%%%%%%%%%%%%%%%%%%%%%%%%%
\section{Background}

\subsection{Prior Knowledge for Perturbation Prediction}

The dominant paradigm for perturbation prediction relies on propagating signals through structured knowledge graphs. GNN-based approaches such as GEARS \citep{roohani2024predicting} and TxPert \citep{wenkel2025txpert} leverage Gene Ontology (GO) \citep{ashburner2000gene} and protein interaction networks \citep{szklarczyk2023string} to share information between perturbations \citep{battaglia2018relational, kipf2016semi}. While effective, these methods operate under a closed-world assumption: if a regulatory relationship is absent from the graph, the model cannot leverage it. They also treat the interactome as static, failing to capture context-specific rewiring across cell types \citep{ideker2012differential}.

An alternative line of work uses LLMs to incorporate biological knowledge without relying on fixed graph structure \cite{phillips2025synthpert}. GenePT \citep{chen2024genept} and GP+LLM \citep{martens2024enhancing} derive gene representations from LLM embeddings, while scGPT \citep{cui2024scgpt} learns representations from tokenized observational single-cell expression profiles. LangPert \citep{martens2025langpert} prompts LLMs directly to identify functionally similar genes for nearest-neighbor prediction. However, as discussed in Section 1, these approaches reflect symmetric co-occurrence in scientific text rather than directed regulatory relationships: they do not distinguish whether $A \rightarrow B$, $B \rightarrow A$, or neither.

MechPert addresses this by treating LLM outputs as noisy hypotheses rather than factual assertions. Rather than relying on a single retrieval, multiple agents independently generate directed regulatory hypotheses, which are filtered through consensus and validated against topological constraints. This decouples co-occurrence from regulatory directionality while retaining the generative flexibility of LLMs.

% In contrast, we propose a shift to dynamic causal reasoning. Unlike fixed graphs, our agentic framework induces context-specific hypotheses from the `open world' of literature \citep{lin2020pubmedbert}. This allows us to predict links for understudied genes despite sparse priors \citep{szklarczyk2023string}, utilizing topological consistency merely as a validity constraint rather than a prerequisite

% \subsection{LLMs for Biological Representation and Reasoning}

% Recent approaches adapt LLMs as \textbf{hypothesis generators} via static embeddings or tokenized sequences, exemplified by \textbf{GenePT} \citep{chen2024genept}, \textbf{scGPT} \citep{cui2024scgpt}, and \textbf{GP+LLM} \citep{martens2024enhancing}. However, these methods suffer from intrinsic \textbf{representation bias}: embeddings trained on distributional co-occurrence ($P(B|A)$) fail to capture regulatory causality ($P(B|\text{do}(A))$) \citep{pearl2009causality, peters2017elements}. This leads to \textbf{structural conflation}, where high lexical similarity \citep{lee2020biobert, beltagy2019scibert} obscures the distinction between upstream drivers and downstream targets, incorrectly implying symmetric regulation \citep{zheng2018dags}.

% To address this, we treat LLM outputs as \textbf{noisy propositions} rather than facts. MechPert acts as a \textbf{reasoning constraint layer}, enforcing consensus and structural consistency to decouple lexical similarity from mechanistic directionality.

\subsection{Experimental Design for Perturbation Screens}

Beyond prediction, selecting which perturbations to perform first is itself a critical design problem. Traditional screens prioritize targets using heuristics such as degree centrality in protein interaction networks \citep{albert2000error, barabasi1999emergence}, but these structural measures do not capture context-specific regulatory logic. Bayesian and active learning approaches \citep{snoek2012practical, settles2009active} offer principled alternatives but require an initial dataset to calibrate uncertainty, making them ineffective in cold-start settings \citep{schein2002methods}. We show that LLM-generated regulatory hypotheses can also guide experimental prioritization, identifying informative anchor perturbations without requiring any prior screening data.

% \subsection{few-shot Prioritization for Exploratory Screening } 
% Traditional large-scale perturbation screens rely on heuristics such as high expression or degree centrality to select targets \citep{albert2000error, barabasi1999emergence}. From the perspective of Active Causal Learning, these structural hubs maximize global connectivity but fail to capture the specific regulatory logic driving cell-state transitions \citep{ he2008active}. While Bayesian Optimization \citep{snoek2012practical} and uncertainty sampling \citep{settles2009active} offer principled frameworks for sequential design, they fundamentally require an initial `seed' dataset to calibrate uncertainty estimates \citep{houlsby2011bayesian}, rendering them ineffective for the \textbf{cold-start problem} \citep{schein2002methods}. We propose a novel framework for \textbf{few-shot Active Design}. By leveraging LLM agents as generative priors, we synthesize literature-derived hypotheses about regulatory drivers \textit{ex ante} \citep{brown2020language, wei2022chain}. This allows us to identify a set of \textbf{Causal Anchors} that maximize expected information gain across the full regulatory network \citep{mackay1992information}, enabling efficient experimental design without requiring any prior screening data.

%%%%%%%%%%%%%%%%%%%%%%%%%%%%%%%%%%%%%%%%%%%%%%%%%%%%%%%%%%
\section{Methods}

\begin{figure}[t]
    \centering
    \includegraphics[width=0.9\textwidth]{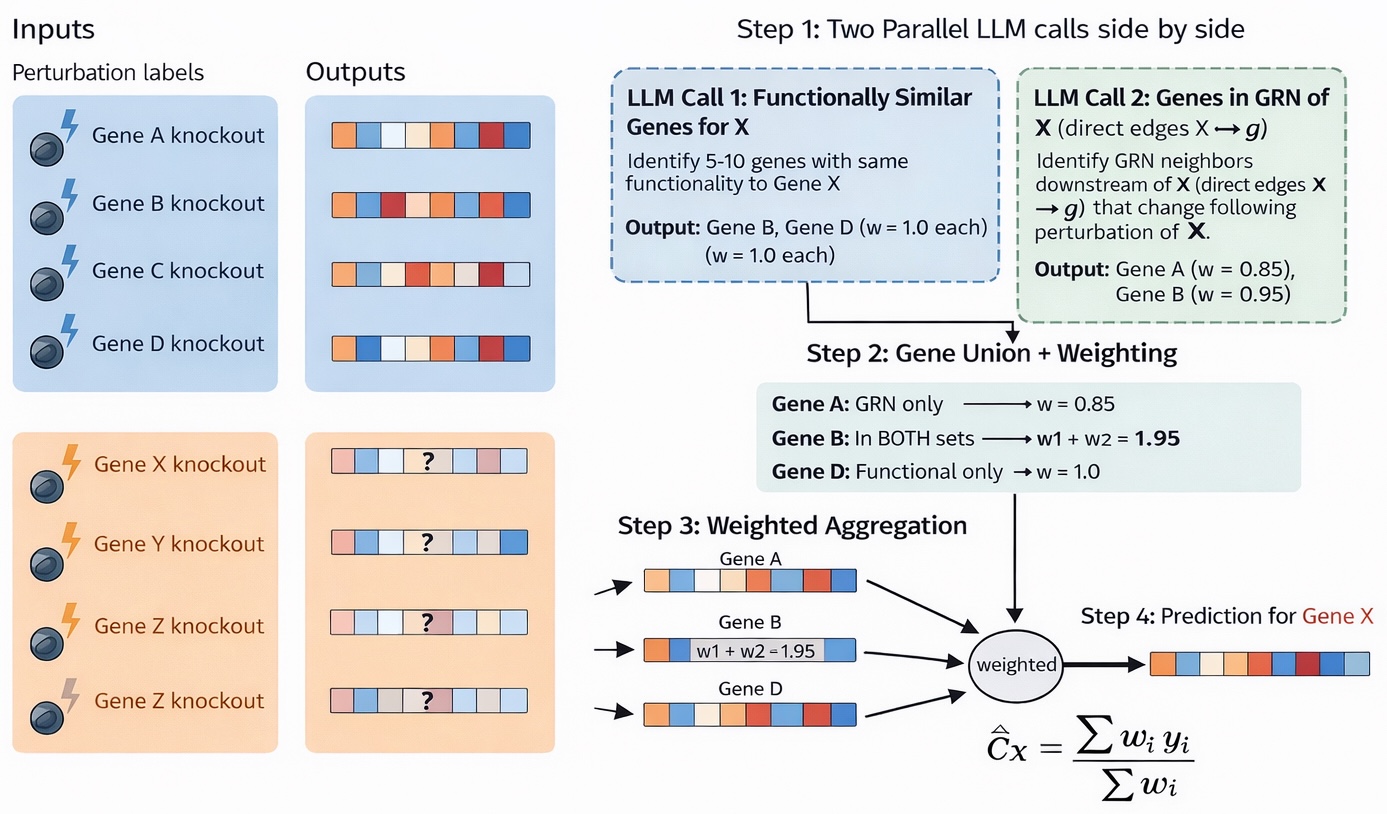}
    \caption{\textbf{MechPert Framework.} Problem setup for predicting gene expression vectors of unseen perturbations.  \textit{MechPert} pipeline: LLMs identify functional similarity and GRN regulators to weight and aggregate training data for few-shot prediction ($\hat{y}_X$). }
    \label{fig:langpert_overview}
\end{figure}

We propose \textsc{MechPert} (Figure \ref{fig:langpert_overview}), a framework that transforms the LLM from a passive retrieval engine into a structured hypothesis generator with directional regulatory constraints. We evaluate this framework on two distinct tasks: (1) \textbf{Few-Shot Perturbation Prediction}, identifying the effect of a specific perturbation; and (2) \textbf{Active Discovery}, identifying which perturbations yield the most information about the system. 

%We propose \textsc{MechPert}, a neuro-symbolic framework that formalizes few-shot perturbation prediction as a probabilistic inference problem over latent regulatory graphs. Our approach treats the LLM not merely as a retrieval engine, but as a generative causal oracle sampled via Monte Carlo consensus. We decompose the prediction task into two stages:

%\begin{enumerate}
 %   \item \textbf{Generative Hypothesis Formation}, where agents propose directed regulatory edges ($u \to v$) based on open-world literature; and
  %  \item \textbf{Topological Consistency Filtering}, where causal confidence is estimated by aggregating these independent reasoning chains.
%\end{enumerate}

\subsection{Few-Shot Perturbation Prediction}

Given a set of $N$ observed perturbation profiles $\{(\mathbf{y}_j, g_j)\}_{j=1}^N$ and a query gene $g_t$ not in the training set, the goal is to predict the transcriptomic response $\hat{\mathbf{y}}_t$. We describe how MechPert constructs this prediction in two steps.

\textbf{Step 1: Dual Hypothesis Generation.} For each query gene $g_t$, we prompt the LLM to generate two distinct candidate sets from the training pool:
\begin{itemize}
\item \textbf{Semantic Set ($\mathcal{N}_{sem}$):} Genes identified by functional similarity to $g_t$ (shared pathways, co-regulation, similar knockout phenotypes). This corresponds to the standard LangPert retrieval strategy \citep{martens2025langpert}.
\item \textbf{Causal Set ($\mathcal{N}_{causal}$):} Genes identified as potential directed regulators of $g_t$ (e.g., upstream transcription factors or direct regulatory targets). For each causal hypothesis $r_i$, the agent also provides a confidence score $c_i \in [0, 1]$ reflecting its certainty in the directed edge $r_i \to g_t$.
\end{itemize}
The causal set is the key addition: by explicitly prompting for directed relationships, we encourage the model to reason about regulatory logic rather than symmetric similarity. To reduce sensitivity to any single LLM call, we execute $K$ independent reasoning chains (we use $K=3$) and retain genes that appear across multiple chains.

\textbf{Step 2: Consensus Aggregation.} The predicted response $\hat{\mathbf{y}}_t \in \mathbb{R}^d$ is computed as a weighted average of the perturbation profiles of the retrieved genes:
\begin{equation}
\hat{\mathbf{y}}_t = \frac{1}{Z} \sum_{j \in \mathcal{N}_{sem} \cup \mathcal{N}_{causal}} w_j \cdot \mathbf{y}_j
\end{equation}
where $Z = \sum_j w_j$. We evaluate two weighting schemes:
\begin{itemize}
\item \textbf{Binary Consensus ($w_j = 1$):} All retrieved genes contribute equally, with consensus arising from the frequency of retrieval across independent chains.
\item \textbf{Confidence-Weighted:} Semantic neighbors receive unit weight ($w_j = 1$), while causal hypotheses are scaled by the agent's reported confidence $c_j$ ($w_j = c_j$).
\end{itemize}
This aggregation is intentionally simple: it treats all regulatory relationships additively, without distinguishing activation from repression. Our goal is to test whether encouraging directed regulatory reasoning alone provides a useful inductive bias in sparse regimes, while acknowledging that sign-aware aggregation remains future work.

\subsection{Active Discovery of Regulatory Anchors}

The second task addresses experimental design: given a budget of $k$ perturbations, which genes should be perturbed first to maximally inform predictions across the full regulatory landscape? Traditional approaches select targets by degree centrality in PPI networks \citep{barabasi1999emergence}, but these structural hubs do not necessarily capture context-specific regulatory logic \citep{he2008active}.

\begin{figure}[t]
    \centering
    \includegraphics[width=0.9\textwidth]{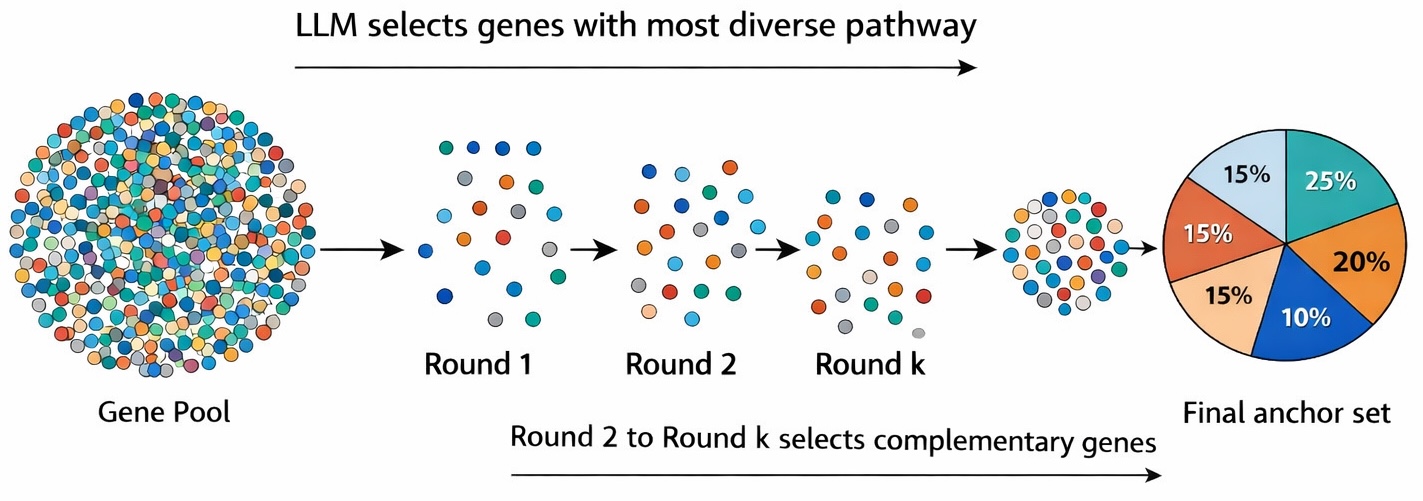}
    \caption{\textbf{Active experiment design.} Iterative hub selection to maximize biological pathway diversity in the final anchor set.}
    \label{fig:active_experiment}
\end{figure}

We use MechPert to identify an informative anchor set $\mathcal{S}$ (Figure \ref{fig:active_experiment}). Analogous to Task I, we prompt LLM agents to identify candidate master regulators for the specific cellular context, aggregating both semantic and causal signals across independent reasoning chains. To encourage diversity across regulatory pathways, selection proceeds iteratively: in each round, genes are selected to complement the pathways already represented in $\mathcal{S}$. The resulting anchor set is then mapped to the physical PPI interactome to evaluate its information propagation potential.

% We extend the framework to \textbf{Active Causal Discovery}: selecting an optimal intervention set $\mathcal{S}$ that maximizes information gain regarding the global regulatory landscape \citep{he2008active}. While traditional heuristics rely on degree centrality in static PPI networks \citep{barabasi1999emergence}, they fail to capture context-specific regulatory logic.

% To overcome this, we prompt the LLM to identify high-value \textbf{``Causal Hubs''} genes predicted to be master regulators of the specific biological context as in Figure \ref{fig:active_experiment}. Analogous to Task I, we aggregate both {semantic} (functional similarity) and {causal} (upstream/downstream driver) signals to rank candidate genes. We select the top-$k$ genes that maximize {Consensus Semantic Centrality}, creating a starting set $\mathcal{S}$ that reflects mechanistic reasoning rather than static topology. These anchors are then mapped to the physical PPI interactome to evaluate their information propagation potential.

%To guide experimental design, we utilize the inferred posterior distribution to identify Semantic Hubs—genes with high expected causal influence ($E[|\text{Children}|]$) across the generated latent graphs. Unlike structural hubs (Degree Centrality), which prioritize static connectivity, Semantic Hubs are selected for their functional regulatory potential. We validate these anchors by measuring their ability to propagate information through the physical interactome (PPI) via graph diffusion, demonstrating that semantically-derived priors maximize information flow in unseen biological contexts.

\subsection{Validation Protocol for Active Discovery}

To rigorously quantify the information content of the anchors $\mathcal{S}$ selected in Task II, we employ a \textit{Surrogate Evaluation Strategy}. Crucially, we decouple \textit{anchor selection} from \textit{inference} to eliminate the confounding variable of the LLM's generative variance. We do \textit{not} use the MechPert engine for this step. Instead, we fix the downstream predictor to be a deterministic \textit{Heat Kernel Interpolator} over the physical PPI network \citep{kondor2002diffusion}.

For any unseen gene $v$, the predicted perturbation effect $\hat{y}_v$ is computed as the weighted average of the anchor effects:
\begin{equation}
    \hat{y}_v = \frac{\sum_{s \in \mathcal{S}} K_{\beta}(s, v) \cdot y_s}{\sum_{s \in \mathcal{S}} K_{\beta}(s, v)}
\end{equation}
where $K_{\beta}(s, v) = [\exp(-\beta \mathbf{L})]_{sv}$ is the heat kernel corresponding to the graph Laplacian $\mathbf{L}$.

By utilizing a fixed, topology-only propagator, we ensure that any performance gain in the \textsc{MechPert} arm is strictly attributable to the \textit{Topological Superiority} of the selected anchors—proving they are better positioned within the functional manifold to capture global variance than standard structural hubs. This evaluation does not measure true experimental utility, but isolates the effect of anchor selection under a fixed inference mechanism.

\section{Results}

\subsection{Structural Priors for Resource-Efficient Models}

We first characterized the limits of LLM reasoning within transcriptomic space. We found that the raw predictions of efficient models, such as Gemini 3 Flash and GPT 5 mini, were frequently challenged by biological signal noise. To bridge this gap, we evaluated three adaptations that anchor semantic retrieval in physical manifolds: (1) \textbf{3+2 Strategy} (Euclidean PPI expansion), (2) \textbf{Topological Harmonizer} (multiscale graph smoothing), and (3) \textbf{Spectral Filtering} (Hadamard-gated Poincaré density). 

These methods act as structural scaffolding, substituting internal model parameters with Protein-Protein Interaction (PPI) and hyperbolic manifold priors. The impact of this scaffolding is most pronounced in low-data regimes: for the \textit{Jurkat} lineage at $N=50$, the \textbf{Topological Harmonizer} improves the Pearson correlation of the  model by \textbf{+24.1\%} ($0.360 \to 0.447$). Strikingly, this augmentation allows the efficient model to surpass the few-shot performance floor of the $100\times$ larger \textit{Pro} backbone ($0.406$). While critical for smaller models, these augmentations provide limited benefit to the largest LLMs, which already possess robust internal representations of these signaling topologies.

\begin{figure*}[t]
\centering
\includegraphics[width=0.9\textwidth]{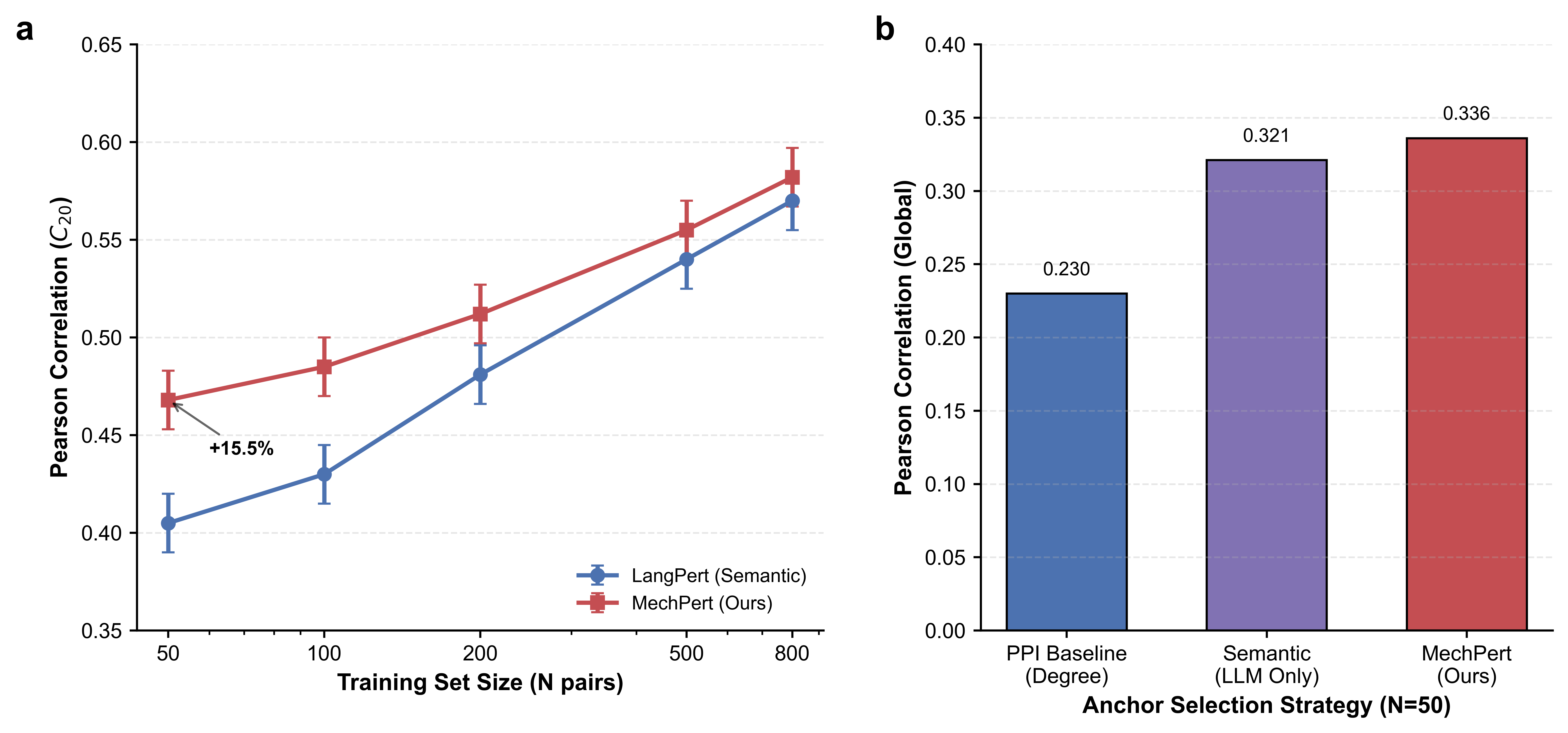}
\caption{\textbf{MECHPERT improves low-data generalization and experimental design.} (a) In Jurkat T-cells, our Causal-Consensus model (red) beats the semantic baseline (blue) by +15.5\% at N=50, showing causal priors help when data are scarce. (b) In K562, choosing N=50 experimental anchors via geometrically adjudicated consensus gives a +46\% gain over PPI-degree heuristics, indicating agentic reasoning finds effective regulatory hubs.}
\label{fig:main_results}
\end{figure*}

\subsection{Causal Priors Improve Sample Efficiency in Few-Shot Regimes}

To test the hypothesis that causal consensus acts as a robust inductive bias when training data is scarce, we evaluated \textsc{MechPert} across four divergent cell lines (K562, RPE1, Jurkat, HepG2). We conducted a Scaling Law Analysis by training on subsampled datasets ranging from $N=50$ to $N=800$ perturbation pairs (Figure \ref{fig:main_results}a). We compare three strategies: (1) \textbf{LangPert} (standard semantic retrieval), (2) \textbf{Binary Consensus} (unweighted voting), and (3) \textbf{\textsc{MechPert}} (confidence-weighted)\cite{martens2025langpert, stoisser2025towards_2}.

As shown in Table \ref{tab:scaling_results}, \textsc{MechPert} consistently outperforms the semantic baseline, particularly in the critical cold-start regime ($N=50$). Aggregated across all cell lines, our confidence-weighted model achieves a +10.4\% relative improvement in Pearson correlation ($C_{20}: 0.528 \to \mathbf{0.583}$) compared to the LangPert baseline.

Crucially, the performance gap is context-dependent. In cell lines with sparse prior knowledge (e.g., Jurkat T-cells), the benefit of causal reasoning is magnified: \textsc{MechPert} improves correlation by +15.5\% ($0.405 \to \mathbf{0.468}$), validating our hypothesis that generative priors effectively substitute for missing experimental data. While Binary Consensus performs comparably to Confidence-Weighted aggregation in some regimes, the weighted approach provides strictly lower variance (Standard Error $\pm 0.031$ vs $\pm 0.035$), suggesting that uncertainty calibration stabilizes the consensus mechanism.

\begin{table}[h]
\centering
\small
\caption{\textbf{Few-Shot Generalization ($C_{20}$ Correlation).} MECHPERT (confidence-weighted) outperforms LangPert and binary consensus on pooled C20 correlation across four cell lines (K562, RPE1, Jurkat, HepG2; Mean ± SEM). Best values per ($N$) are highlighted, and MECHPERT shows a notable +10.4\% gain in the low-data setting ($N=50$).}
\label{tab:scaling_results}
\begin{tabular}{lcccc}
\toprule
\textbf{Training Size ($N$)} & \textbf{LangPert (Sem)} & \textbf{Binary Consensus} & \textbf{\textsc{MechPert} (Conf)} & \textbf{Rel. Improv.} \\
\midrule
$N=50$  & $0.528 \pm 0.032$ & $0.553 \pm 0.031$ & $\mathbf{0.583}\pm 0.031$ & \textbf{+10.4\%} \\
$N=100$ & $0.529 \pm 0.030$ & $0.540 \pm 0.029$ & $\mathbf{0.551} \pm 0.029$ & \textbf{+4.1\%} \\
$N=200$ & $0.558 \pm 0.029$ & $0.569 \pm 0.029$ & $\mathbf{0.581} \pm 0.029$ & \textbf{+4.1\%} \\
$N=500$ & $0.583 \pm 0.028$ & $\mathbf{0.606} \pm 0.028$ & $0.589 \pm 0.028$ & +3.9\% \\
$N=800$ & $0.605 \pm 0.027$ & $\mathbf{0.634} \pm 0.026$ & $0.620 \pm 0.026$ & +4.8\% \\
\bottomrule
\end{tabular}
\end{table}

\textbf{Agentic Selection Outperforms Structural Heuristics.}
To evaluate the utility of agentic reasoning for experimental design, we compared four anchor selection strategies for $N=50$ in Table \ref{tab:active_discovery}: (1) Random Uniform (Maximum Entropy), (2) PPI Degree Centrality (Structural Baseline), (3) Semantic Importance (LLM Only), and (4) \textsc{MechPert} (Consensus).

Our results reveal a distinct regime of advantage. In well-characterized lineages where the structural prior is dense (K562, RPE1), \textsc{MechPert} identifies anchors that maximize information flow relative to structural centrality. Specifically, in K562, our consensus strategy yields a +46\% improvement over the standard PPI Degree baseline ($0.336$ vs. $0.230$), confirming that semantic hubs are topologically superior to naive structural bottlenecks for signal propagation.

However, we observe that performance is strictly conditioned on the fidelity of the biophysical prior. In contexts with sparser interactomes (HepG2) or highly dynamic signaling (Jurkat), the synergy between text and graph breaks down ($0.181$ vs. $0.247$), and simple Random Uniform sampling—which maximizes global coverage rather than local mechanism—remains the most robust strategy ($0.261$). This validates the hypothesis that agentic reasoning acts as a structural amplifier: it can enhance the utility of well-characterized biological networks but cannot invent topology where physical ground truth is missing.

\begin{table}[h]
\centering
\small
\caption{\textbf{Active Experimental Design Performance.} Pearson correlation of global genome prediction using $N=50$ anchors selected by different strategies. \textsc{MechPert} provides the best targeted strategy in well-characterized lineages (K562, RPE1), beating standard PPI centrality.}
\label{tab:active_discovery}
\begin{tabular}{lcccc}
\toprule
\textbf{Cell Line} & \textbf{Random (Reference)} & \textbf{PPI Baseline} & \textbf{Semantic (LLM)} & \textbf{\textsc{MechPert} (Ours)} \\
\midrule
K562   & $0.404$ & $0.230$ & $0.321$ & $\mathbf{0.536}$ \\
RPE1   & $0.
580$ & $0.592$ & $0.612$ & $\mathbf{0.614}$ \\

Jurkat & $\mathbf{0.367}$ & $0.315$ & $0.316$ & $0.260$ \\
HepG2  & $\mathbf{0.261}$ & $0.247$ & $0.199$ & $0.181$ \\
\bottomrule
\end{tabular}
\end{table}

Taken together, these results establish a dual advantage for consensus-driven reasoning. First, it enables robust few-shot generalization by imposing severe inductive biases when training data is scarce ($N=50$). Second, it transforms the experimental design process itself, allowing agents to autonomously identify the most informative causal interventions before a single experiment is run. This moves toward integrating inference and experimental prioritization, suggesting that LLM agents may assist in navigating combinatorial biological design spaces.

\subsection{Ablation Study: Contribution of Consensus Components}

To rigorously isolate the contribution of consensus aggregation and confidence calibration, we performed an ablation study in the low-data regime ($N=50$) on the challenging Jurkat cell line (Table \ref{tab:ablation}).

Comparing the breakdown of performance gains reveals a clear hierarchy of inductive biases. The transition from Single-Agent Retrieval (Baseline) to Multi-Agent Binary Consensus yields the most significant lift ($+10.6\%$), suggesting that the simple frequency of hypothesis generation across independent reasoning chains is a powerful filter for stochastic hallucination. However, incorporating the agent's explicit uncertainty estimate via Confidence Weighting provides a critical second-order refinement, boosting performance by an additional $+4.9\%$ (Total $+15.5\%$). This confirms that while ensemble consistency establishes robustness, calibration is necessary to achieve state-of-the-art accuracy in sparse data regimes.

\begin{table}[h]
\small
\centering
\caption{\textbf{Ablation Study of Consensus Components ($N=50$).} Jurkat shows most performance gain comes from the consensus aggregation (frequency), with confidence weighting adding further refinement.}
\label{tab:ablation}
\begin{center}
\resizebox{0.95\columnwidth}{!}{%
\begin{tabular}{llcc}
\toprule
\textbf{Model Variant} & \textbf{Aggregation Logic} & \textbf{Mean $C_{20}$ (Jurkat)} & \textbf{\% Gain} \\
\midrule
LangPert (Baseline) & None (Single Retrieval) & $0.405$ & -- \\
Binary Consensus & Unweighted Voting ($w_i=1$) & $0.448$ & $+10.6\%$ \\
\textbf{\textsc{MechPert} (Full)} & \textbf{Confidence Weighted ($w_i=c_i$)} & $\mathbf{0.468}$ & $\mathbf{+15.5\%}$ \\
\bottomrule
\end{tabular}
}
\end{center}
\end{table}

\subsection{Case Study: Disambiguating Stress Pathways via Mechanistic Reasoning}

To demonstrate the interpretability of our approach, we examined the prediction of \textbf{NVL} (Nuclear VCP-Like), a critical AAA-ATPase involved in ribosome biogenesis, within the HepG2 liver carcinoma line. This gene presents a challenging test case due to its high lexical similarity to its paralog VCP, which operates in a fundamentally different cellular compartment.

\textbf{Representation Failure due to Lexical Ambiguity.}
The semantic baseline, relying on distributional co-occurrence statistics, exhibited \textbf{lexical conflation}: it failed to distinguish \textit{NVL} from its paralog \textit{VCP} (Valosin-Containing Protein) due to shared nomenclature and high textual co-occurrence in the training corpus. Consequently, it retrieved functional neighbors associated with the \textit{Ubiquitin-Proteasome System} and \textit{ER-Associated Degradation (ERAD)}. This led to the prediction of an \textbf{Unfolded Protein Response (UPR)} signature (characterized by \textit{ATF4, CHOP, XBP1}), which is biologically incorrect for an NVL knockout. This representational failure resulted in a strong negative correlation with the ground truth ($C_{20} = -0.67$), indicating that the model predicted the \textit{opposite} of the true transcriptomic response.

\textbf{Mechanistic Disambiguation via Causal Consensus.}
In contrast, our \textsc{MechPert} agent explicitly adjudicated the regulatory mechanism by leveraging \textbf{subcellular context}. The consensus reasoning correctly identified that while NVL shares a protein family with VCP, it specifically functions in the \textbf{nucleolus} for \textit{pre-60S ribosomal subunit release}, rather than in the ER for protein quality control.
\begin{enumerate}
    \item \textbf{Mechanism Inferred:} The agent correctly identified that disrupting NVL causes \textbf{Nucleolar Stress}, leading to the release of ribosomal proteins (e.g., \textit{RPL5, RPL11}) into the nucleoplasm, where they sequester MDM2.
    \item \textbf{Outcome Predicted:} Instead of a UPR signature, the agent predicted a stabilization of \textbf{p53} and the massive upregulation of its downstream targets (\textit{CDKN1A/p21, MDM2, BTG2}).
\end{enumerate}

By correctly mapping the perturbation to the \textbf{p53-dependent Ribosomal Stress Pathway} rather than the \textbf{Proteotoxic Stress Pathway}, the Consensus model achieved a strong positive correlation with the ground truth ($C_{20} = 0.74$). This represents a $\Delta C_{20} = 1.41$ swing from failure to success, attributable solely to the difference in reasoning architecture. This case demonstrates that the ability to disentangle distinct stress mechanisms illustrates that disentangling stress mechanisms may improve perturbation prediction.

\section{Discussion}
\label{sec:discussion}

Our results reveal a fundamental limitation of treating large language models solely as retrieval engines: semantic proximity in the literature is not isomorphic to functional causality in the cell. While standard LLMs excel at capturing distributional co-occurrence, they inherently struggle to represent the directed, asymmetric nature of gene regulatory networks.

% \subsection{Consensus as a Proxy for Causal Probability}
The consistent performance gains observed in the low-data regime ($N \leq 100$) demonstrate that multi-agent consensus provides a robust inductive bias when training data is scarce. This positions \textsc{MechPert} as a highly effective \textbf{`few-shot Cold-Start' engine}. While data-rich regimes ($N \ge 800$) allow supervised models to bridge the performance gap, our approach provides the critical prior knowledge necessary for the initial discovery phase of understudied genes where experimental data is non-existent.

Crucially, hypothesis frequency may act as a heuristic proxy for regulatory plausibility. By aggregating votes from multiple LLM agents, we implicitly filter stochastic hallucinations in favor of reproducible causal signals \cite{stoisser2025towards}. This provides a natural explanation for the method's dominance in sparse regimes: the consensus mechanism appears to provide an inductive bias that reduces reliance on purely associative neighbors that lack consistent mechanistic support across diverse reasoning paths.

\section{Limitations}
MechPert is evaluated on four human cancer cell lines using single-gene CRISPRi perturbations; generalization to primary cells, gene knockouts, and combinatorial or temporal perturbations remains untested. The method is most effective when partial regulatory structure is available but incomplete. In poorly characterized lineages such as HepG2 and Jurkat, LLM-guided hypothesis generation can underperform random sampling, likely reflecting publication and annotation biases in pretraining corpora. Additionally, since the evaluation benchmark \cite{replogle2022mapping} is a high-profile publication likely present in the model's training data, future work should evaluate on perturbations from post-training-cutoff publications to rule out memorization effects.

Additionally, the validation of MechPert has been conducted primarily in controlled cell line models with specific genetic perturbations. Its effectiveness and robustness in complex, heterogeneous, and multi-cellular biological systems remain untested. Extensive validation in primary tissues, disease models, or in vivo settings is necessary to confirm its practical utility in real-world biological and therapeutic contexts \cite{ott2025identification}.

The framework relies on consensus across multiple LLM agents rather than calibrated confidence estimates. We observe that majority voting performs comparably to or better than confidence-weighted aggregation, indicating that model-reported confidence is not reliably aligned with biological plausibility in this setting.

Finally, causal language in this work refers to directed regulatory hypotheses rather than formal causal identification. MechPert does not estimate interventional effect sizes or recover a uniquely identified causal graph; its outputs should be interpreted as mechanistically informed hypotheses for experimental prioritization. Future work could incorporate causal effect estimation methods to strengthen these claims.

\section*{Acknowledgments}
We acknowledge the use of LLMs to assist with text editing, distinctness, grammatical refinement and diagram editing. All scientific claims and the final text were reviewed and verified by the authors.
\bibliographystyle{plain} % Choose the desired bibliography style
\bibliography{bib} % Ensure 'bib.bib' exists in the same directory as this LaTeX file

\appendix

\renewcommand{\thetable}{\thesection.\arabic{table}}
\renewcommand{\thefigure}{\thesection.\arabic{figure}}
\renewcommand{\theequation}{\thesection.\arabic{equation}}

\section{Architectural Variants and Adaptations}\label{app:Archit}
To evaluate the scaling of different inductive priors, we compare the baseline model against three distinct architectural adaptations. These adaptations were developed primarily for smaller LLM models (Gemini 3 Flash, gpt 5 mini) to augment their ability to retrieve and process sparse biological signals. Preliminary ablations showed that larger LLMs did not benefit significantly from these specific topological expansion strategies due to their superior internal reasoning capabilities; consequently, these methods were not applied to the larger model benchmarks.

\subsection{Geometric Data-Augmented Retrieval (The 3+2 Strategy)}

While the standard LangPert baseline relies on a Large Language Model (LLM) to retrieve a flexible set of 5, the 3+2 strategy introduces a geometric prior to stabilize retrieval for smaller models. This approach leverages a hybrid reasoning-manifold pipeline consisting of the following steps:

\begin{enumerate}
    \item \textbf{Expert Seed Retrieval:} The model first queries the LLM to identify exactly three ``Expert Seeds'' from the training pool. These genes are selected based on the highest perceived functional similarity (e.g., shared biological pathways and co-regulation) to the target gene $g_t$.
    
    \item \textbf{Euclidean Manifold Projection:} The identified seeds are projected into a 50-dimensional Euclidean embedding space ($\mathbb{R}^{50}$). These embeddings are pre-trained using Node2Vec on the STRING Protein-Protein Interaction (PPI) interactome, ensuring that proximity in the latent space corresponds to structural and functional relatedness in the proteome.
    
    \item \textbf{Neighborhood Completion:} We calculate the Euclidean centroid $C$ of the three expert seeds. To ground the prediction in the local topological density of the manifold, the model identifies the two nearest neighbors (``geometric neighbors'') to this centroid from the training pool:
    \begin{equation}
        C = \frac{1}{3} \sum_{i \in \text{Seeds}} \mathbf{e}_i
    \end{equation}
    where $\mathbf{e}_i$ represents the embedding vector of the $i$-th seed.
    
    \item \textbf{Robust Aggregation:} The final predicted perturbation profile $\hat{y}_t$ is computed as a simple average of the perturbation vectors from the ensemble of five genes (3 semantic seeds plus 2 geometric neighbors):
    \begin{equation}
        \hat{y}_t = \frac{1}{5} \left( \sum_{i \in \text{Seeds}} y_{i} + \sum_{j \in \text{neighbors}} y_{j} \right)
    \end{equation}
\end{enumerate}

By delegating the neighborhood completion to the Euclidean PPI manifold, the 3+2 strategy ensures that the functional context is anchored to the physical interactome, effectively ``rescuing'' the prediction when the LLM's primary retrieval is sparse or uncertain.

\subsubsection{Topological Manifold Harmonizer}

Building upon the discrete retrieval of the \textit{LangPert} baseline and the Euclidean-based \textit{Geometric Manifold Retrieval} (the 3+2 strategy), the \textbf{Topological Manifold Harmonizer} performs a continuous signal broadcast across the physical interactome. While standard \textit{LangPert} relies exclusively on a small set of agent-selected semantic seeds, and the 3+2 strategy provides a discrete expansion in flat Euclidean space, the Harmonizer variant utilizes graph-theoretic diffusion and hyperbolic geometry to aggregate a broader neighborhood of the top 50 most reachable genes. This process consists of two primary stages:

\begin{enumerate}
    \item \textbf{Topological Signal Propagation:} 
    Using the \textsf{networkx} Python package, the model performs a Personalized PageRank (PPR) traversal on the STRING PPI interactome ($G_{\text{PPI}}$). In this implementation, the three LLM-selected seeds are treated as starting points for a diffusion process. We initialize a personalization vector $\mathbf{p}$ with uniform weight across the seeds and compute the steady-state reachability distribution using a damping factor $\alpha = 0.85$:
    \begin{equation}
        \mathbf{pr} = \alpha \mathbf{A} \mathbf{pr} + (1 - \alpha) \mathbf{p}
    \end{equation}
    where $\mathbf{A}$ represents the transition matrix of the interactome. This traversal identifies the top 50 genes most topologically relevant to the original expert seeds.

    \item \textbf{Hyperbolic Density Weighting:} 
    To refine the expanded neighborhood, we utilize hierarchical embeddings in a 100-dimensional Poincaré ball, implemented via the \textsf{gensim} package's \texttt{PoincareModel}. For each identified neighbor $j$, we calculate its hyperbolic distance $d_{\mathbb{H}}$ to the Einstein midpoint $C$ of the original semantic seeds:
    \begin{equation}
        w_j = \exp \left( -\frac{d_{\mathbb{H}}(\mathbf{z}_j, C)^2}{2\sigma^2} \right)
    \end{equation}
    where $\mathbf{z}_j$ is the coordinate vector on the Poincaré manifold. To maintain high specificity, the bandwidth $\sigma$ is set dynamically based on the $20^{\text{th}}$ percentile of local distances. 
\end{enumerate}

\textbf{Relationship to Previous Strategies:} Unlike the \textit{LangPert} baseline which assumes independent retrieval, or the \textit{3+2 strategy} which uses flat Euclidean geometry to ``rescue'' sparse sets with a few extra genes, the \textbf{Topological Manifold Harmonizer} utilizes \textsf{gensim}'s hyperbolic coordinates to model the latent hierarchy of biological systems. By transitioning from discrete neighbor selection to continuous manifold smoothing via \textsf{networkx}, this approach allows the model to identify relevant functional neighbors that may be semantically distant to an LLM but are structurally essential to the target perturbation.

\subsubsection{Spectral Manifold Harmonizer}

The \textbf{Spectral Manifold Harmonizer} represents our most rigorous adaptation, designed to isolate the highest-fidelity functional signals within the proteome. While the \textit{Topological Manifold Harmonizer} performs a standard smoothing across reachable nodes, the Spectral variant treats the manifold as a conductive surface where signal strength is modulated by local hierarchical density. This method integrates topological heat and manifold conductivity through a multiplicative gating process:

\begin{enumerate}
    \item \textbf{Topological Heat Diffusion:} 
    Similar to the Harmonizer variant, we utilize the \textsf{networkx} library to simulate a diffusion process on the STRING PPI interactome ($G_{\text{PPI}}$). We compute a Personalized PageRank distribution $\mathbf{pr}$ starting from the semantic seeds. This distribution, which we define as the ``Topological Heat,'' represents the global reachability of all candidate genes from the expert reasoning of the LLM.
    
    \item \textbf{Hyperbolic Conductivity Gating:} 
    In parallel, we calculate the hierarchical relevance of each neighbor using coordinates from the \textsf{gensim} \textit{PoincareModel}. For each reachable gene $j$, we compute a manifold density weight $w_{\text{geo}, j}$ based on its hyperbolic distance to the semantic center:
    \begin{equation}
        w_{\text{geo}, j} = \exp\left(-\frac{d_{\mathbb{H}}(\mathbf{z}_j, C)^2}{2\sigma_{\text{spec}}^2}\right)
    \end{equation}
    where $d_{\mathbb{H}}$ is the Poincaré metric. In the Spectral variant, the distance bandwidth $\sigma_{\text{spec}}$ is constrained to the $15^{\text{th}}$ percentile (rather than 20th) to enforce a tighter structural constraint.

    \item \textbf{Spectral Weight Product:} 
    The final weighting $w_{\text{final}, j}$ for each gene in the ensemble is calculated as the Hadamard (element-wise) product of the topological heat and the manifold conductivity:
    \begin{equation}
        w_{\text{final}, j} = \text{pr}_j \times w_{\text{geo}, j}
    \end{equation}
    By multiplying these scores, the model applies a physical ``gate'': a gene must be both highly reachable in the interactome hierarchy \textit{and} reside in the correct hierarchical branch of the manifold to influence the prediction.
\end{enumerate}

\textbf{Relationship to Previous Strategies:} The \textbf{Spectral Manifold Harmonizer} differs from the standard \textit{Harmonizer} by transitioning from additive signal smoothing to multiplicative signal gating. While the previous variants allow for broader context, the Spectral adaptation is designed for high-fidelity discovery, effectively pruning any topological neighbors that lack hierarchical support in the Poincaré ball. This dual-filter approach ensures that the resulting perturbation vector is maximally consistent with both the agent's semantic logic and the latent tree-like structure of the proteome.

\section{Results}

\subsection{Scaling Analysis of Agentic Consensus}

The results in Table~\ref{tab:pro_results_final_clean} demonstrate that the \textit{LangPert} consensus framework consistently enhances the predictive fidelity of the \textsf{Gemini 3 Pro} model across all scaling regimes. In well-characterized cell lines such as \textbf{RPE1} and \textbf{K562}, we observe a clear hierarchy: \textbf{Weighted Consensus} typically provides the most significant gains in low-data environments ($N \leq 200$) by leveraging the model's self-reported confidence to prioritize high-fidelity associations. As the training pool expands to $N=800$, the \textbf{Binary Consensus} variant takes the lead, achieving peak correlations such as $0.8734$ in RPE1. This shift indicates that frequency-aware multiset aggregation becomes a superior noise-filter as the candidate pool grows. Even in the more challenging \textbf{Jurkat} and \textbf{HepG2} cell lines, these consensus mechanisms successfully stabilize predictions, showing that the aggregation of independent reasoning traces allows the model to self-correct and maintain high structural alignment within complex regulatory manifolds.

\begin{table}[ht]
\centering
\caption{Main Paper Performance Comparison ($C_{20}$ Pearson Correlation $\pm$ SEM) using the \textsf{Gemini 3 Pro} backbone. We evaluate the impact of agentic consensus strategies on prediction fidelity across five training pool sizes ($N$). \textit{LangPert} represents the standard single-agentic retrieval. \textit{Binary Consensus} implements a frequency-aware multiset aggregation across multiple independent reasoning traces. \textit{Weighted Consensus} utilizes the LLM's self-reported confidence scores to weight the contribution of each identified regulator. Maximum values for each row are bolded.}
\label{tab:pro_results_final_clean}
\resizebox{\textwidth}{!}{%
\begin{tabular}{lcccc}
\hline
\textbf{Cell Line} & \textbf{$N$} & \textbf{LangPert} & \textbf{Binary Consensus} & \textbf{Weighted Consensus} \\ \hline
\textbf{RPE1} & 50  & $0.8142 \pm 0.0486$ & $0.8170 \pm 0.0449$ & $\mathbf{0.8227 \pm 0.0456}$ \\
 & 100 & $0.8098 \pm 0.0471$ & $0.8013 \pm 0.0500$ & $\mathbf{0.8116 \pm 0.0483}$ \\
 & 200 & $0.7756 \pm 0.0601$ & $0.7936 \pm 0.0604$ & $\mathbf{0.8003 \pm 0.0598}$ \\
 & 500 & $0.8138 \pm 0.0471$ & $\mathbf{0.8246 \pm 0.0494}$ & $0.8172 \pm 0.0509$ \\
 & 800 & $0.7681 \pm 0.0767$ & $\mathbf{0.8734 \pm 0.0277}$ & $0.8176 \pm 0.0518$ \\ \hline
\textbf{K562} & 50  & $0.5696 \pm 0.0704$ & $0.6063 \pm 0.0607$ & $\mathbf{0.6139 \pm 0.0655}$ \\
 & 100 & $0.6142 \pm 0.0692$ & $0.6234 \pm 0.0670$ & $\mathbf{0.6659 \pm 0.0638}$ \\
 & 200 & $0.6410 \pm 0.0745$ & $0.6475 \pm 0.0769$ & $\mathbf{0.6725 \pm 0.0673}$ \\
 & 500 & $0.6773 \pm 0.0753$ & $\mathbf{0.7048 \pm 0.0689}$ & $0.6725 \pm 0.0750$ \\
 & 800 & $0.7025 \pm 0.0678$ & $\mathbf{0.7203 \pm 0.0720}$ & $0.7120 \pm 0.0679$ \\ \hline
\textbf{Jurkat} & 50  & $0.4057 \pm 0.0888$ & $\mathbf{0.5270 \pm 0.0740}$ & $0.4685 \pm 0.0800$ \\
 & 100 & $0.3801 \pm 0.1054$ & $0.4128 \pm 0.1010$ & $\mathbf{0.4400 \pm 0.0966}$ \\
 & 200 & $0.3777 \pm 0.1086$ & $0.3721 \pm 0.1129$ & $\mathbf{0.4077 \pm 0.1136}$ \\
 & 500 & $0.4293 \pm 0.1149$ & $0.4266 \pm 0.1130$ & $\mathbf{0.4336 \pm 0.1186}$ \\
 & 800 & $0.4424 \pm 0.1191$ & $\mathbf{0.4473 \pm 0.1167}$ & $0.4202 \pm 0.1221$ \\ \hline
\textbf{HepG2} & 50  & $\mathbf{0.2930 \pm 0.0990}$ & $0.2696 \pm 0.1013$ & $0.2532 \pm 0.1048$ \\
 & 100 & $0.5498 \pm 0.0679$ & $0.5380 \pm 0.0634$ & $\mathbf{0.5529 \pm 0.0648}$ \\
 & 200 & $0.4366 \pm 0.0968$ & $\mathbf{0.4620 \pm 0.0950}$ & $0.4460 \pm 0.0980$ \\
 & 500 & $0.4019 \pm 0.0961$ & $\mathbf{0.4691 \pm 0.0869}$ & $0.4312 \pm 0.0918$ \\
 & 800 & $0.5083 \pm 0.0973$ & $0.4946 \pm 0.0857$ & $\mathbf{0.5294 \pm 0.0916}$ \\ \hline
\end{tabular}%
}
\end{table}

\subsection{Augmenting Small Model Capabilities via Inductive Priors}

The scaling results in Table~\ref{tab:c20_results} demonstrate that incorporating structural priors significantly enhances the predictive accuracy of smaller language models. Across all cell lines, we observe a consistent ``topological rescue'' effect: the \textbf{Harmonizer} variant outperforms the baseline in nearly all low-to-mid data regimes, with its most pronounced impact in the \textit{Jurkat} lineage where it improves the correlation by nearly 25\% at $N=50$ ($0.4721$ vs $0.3826$). As data density increases to $N=800$, the optimal strategy shifts toward the \textbf{Spectral Harmonizer} and the \textbf{3+2 Strategy}, which provide the highest fidelity for well-characterized networks like \textit{K562} and the challenging \textit{HepG2} line. Collectively, these results show that while performance generally improves with training size $N$, offloading topological reasoning to a physical manifold allows efficient models to match the performance of significantly larger backbones.

\begin{table}[ht]
\centering
\caption{Scaling Performance Comparison ($C_{20}$ Pearson Correlation $\pm$ SEM) using the \textsf{Gemini 3 Flash} backbone. We evaluate the impact of different inductive priors on the predictive fidelity of smaller language models across five training pool sizes ($N$). \textit{Base LangPert} represents the standard single-agentic retrieval. The \textit{3+2 Strategy} utilizes a discrete Euclidean manifold expansion (3 expert seeds + 2 nearest neighbors). The \textit{Harmonizer} variant implements continuous topological smoothing using Personalized PageRank on the STRING PPI network combined with Poincaré hyperbolic weighting. The \textit{Spectral} variant utilizes a multiplicative Hadamard gate between topological reachability and manifold conductivity to isolate high-fidelity regulatory signals. Maximum values for each configuration (row) are bolded.}
\label{tab:c20_results}
\resizebox{\textwidth}{!}{%
\begin{tabular}{l|c|cccc}
\hline
\textbf{Cell Line} & \textbf{$N$} & \textbf{ LangPert} & \textbf{3+2 Strategy} & \textbf{Harmonizer} & \textbf{Spectral} \\ \hline
\multirow{5}{*}{\textbf{K562}} 
 & 50  & $0.5535 \pm 0.0375$ & $0.5531 \pm 0.0377$ & $\mathbf{0.5898} \pm \mathbf{0.0291}$ & $0.5381 \pm 0.0374$ \\
 & 100 & $0.5732 \pm 0.0397$ & $0.5737 \pm 0.0394$ & $\mathbf{0.6124} \pm \mathbf{0.0318}$ & $0.5320 \pm 0.0406$ \\
 & 200 & $0.6132 \pm 0.0359$ & $0.6112 \pm 0.0358$ & $\mathbf{0.6334} \pm \mathbf{0.0314}$ & $0.5873 \pm 0.0375$ \\
 & 500 & $0.6525 \pm 0.0351$ & $0.6571 \pm 0.0351$ & $0.6458 \pm 0.0329$ & $\mathbf{0.6685} \pm \mathbf{0.0344}$ \\
 & 800 & $\mathbf{0.6673} \pm \mathbf{0.0355}$ & $0.6665 \pm 0.0353$ & $0.6594 \pm 0.0341$ & $0.6653 \pm 0.0374$ \\ \hline
\multirow{5}{*}{\textbf{RPE1}} 
 & 50  & $0.7520 \pm 0.0291$ & $0.7519 \pm 0.0292$ & $\mathbf{0.7686} \pm \mathbf{0.0267}$ & $0.7416 \pm 0.0288$ \\
 & 100 & $0.7418 \pm 0.0300$ & $0.7419 \pm 0.0300$ & $\mathbf{0.7752} \pm \mathbf{0.0260}$ & $0.7396 \pm 0.0296$ \\
 & 200 & $0.7484 \pm 0.0286$ & $0.7471 \pm 0.0287$ & $\mathbf{0.7789} \pm \mathbf{0.0263}$ & $0.7476 \pm 0.0280$ \\
 & 500 & $0.7816 \pm 0.0266$ & $0.7816 \pm 0.0266$ & $\mathbf{0.7932} \pm \mathbf{0.0259}$ & $0.7845 \pm 0.0257$ \\
 & 800 & $0.8001 \pm 0.0246$ & $\mathbf{0.8020} \pm \mathbf{0.0244}$ & $0.7933 \pm 0.0261$ & $0.7799 \pm 0.0279$ \\ \hline
\multirow{5}{*}{\textbf{Jurkat}} 
 & 50  & $0.3826 \pm 0.0489$ & $0.3832 \pm 0.0488$ & $\mathbf{0.4721} \pm \mathbf{0.0509}$ & $0.3926 \pm 0.0493$ \\
 & 100 & $0.4268 \pm 0.0513$ & $0.4272 \pm 0.0513$ & $\mathbf{0.4863} \pm \mathbf{0.0508}$ & $0.4075 \pm 0.0506$ \\
 & 200 & $0.4381 \pm 0.0496$ & $0.4425 \pm 0.0497$ & $\mathbf{0.4723} \pm \mathbf{0.0522}$ & $0.4159 \pm 0.0486$ \\
 & 500 & $0.4492 \pm 0.0475$ & $0.4503 \pm 0.0471$ & $\mathbf{0.4661} \pm \mathbf{0.0552}$ & $0.3835 \pm 0.0519$ \\
 & 800 & $0.4432 \pm 0.0494$ & $0.4530 \pm 0.0482$ & $\mathbf{0.4690} \pm \mathbf{0.0560}$ & $0.4262 \pm 0.0509$ \\ \hline
\multirow{5}{*}{\textbf{HepG2}} 
 & 50  & $0.1788 \pm 0.0465$ & $0.1788 \pm 0.0464$ & $\mathbf{0.1834} \pm \mathbf{0.0565}$ & $0.1374 \pm 0.0476$ \\
 & 100 & $0.2616 \pm 0.0505$ & $\mathbf{0.2633} \pm \mathbf{0.0499}$ & $0.2091 \pm 0.0576$ & $0.2615 \pm 0.0482$ \\
 & 200 & $0.2313 \pm 0.0532$ & $0.2189 \pm 0.0536$ & $0.2085 \pm 0.0577$ & $\mathbf{0.2397} \pm \mathbf{0.0551}$ \\
 & 500 & $0.2896 \pm 0.0502$ & $\mathbf{0.2997} \pm \mathbf{0.0496}$ & $0.2248 \pm 0.0551$ & $0.2525 \pm 0.0516$ \\
 & 800 & $\mathbf{0.3205} \pm \mathbf{0.0506}$ & $0.3203 \pm 0.0510$ & $0.2570 \pm 0.0530$ & $0.2674 \pm 0.0543$ \\ \hline
\end{tabular}%
}
\end{table}

\section{Prompt Templates}
\label{app:prompts}

This section details the exact instruction sets used for both experimental tasks. To ensure high-fidelity biological reasoning, all prompts follow a \textit{Chain-of-Thought} (CoT) protocol requiring the model to articulate mechanistic justifications before returning structured outputs.

%------------------------------------------------------------------------------
\subsection{Task 1: Perturbation Prediction}
%------------------------------------------------------------------------------

\subsubsection{Baseline (LangPert): Functional Similarity Retrieval}

\begin{tcolorbox}[colback=blue!5!white, colframe=blue!75!black, title=Baseline System Prompt]
You are a Lead Computational Biologist. Your task is to perform an \textit{analogous function mapping} for gene perturbations. You must ground your similarity assessments in high-confidence mechanistic evidence, prioritizing genes that occupy identical transcriptomic manifolds or pathway positions. Respond strictly in valid JSON.
\end{tcolorbox}

\begin{tcolorbox}[colback=green!5!white, colframe=green!75!black, title=Baseline User Prompt]
\textbf{Instruction:} Identify \texttt{\{k\_range\}} functional neighbors for the target gene \texttt{\{gene\}}.

\textbf{Available Candidates:} \texttt{\{list\_of\_genes\}}

\textbf{Chain-of-Thought Protocol:}
\begin{enumerate}
    \item \textbf{Profile Analysis:} Define the metabolic/signaling role of \texttt{\{gene\}} in the \texttt{\{context\}}.
    \item \textbf{Pathway Alignment:} For each top-ranked candidate, identify the specific shared interaction (e.g., membership in the same protein complex or signaling cascade).
    \item \textbf{Adjudication:} Rank neighbors based on the depth of the mechanistic link.
\end{enumerate}

\textbf{Output Format (JSON):} \texttt{\{"reasoning": \{... \}, "kNN": ["G1", "G2", ...]\}}
\end{tcolorbox}

\subsubsection{MechPert (Ours): Mechanistic Grounding via Regulatory Anchors}

\begin{tcolorbox}[colback=red!5!white, colframe=red!75!black, title=MechPert System Prompt]
You are a Molecular Systems Geneticist. Your objective is not general similarity, but the identification of the \textit{Mechanical Drivers} of a gene's perturbation profile. Priority: Transcription Factors (TFs), downstream targets, or obligate complex partners.
\end{tcolorbox}

\begin{tcolorbox}[colback=white, colframe=black, title=MechPert User Prompt]
\textbf{Task:} Map the regulatory hierarchy of \texttt{\{gene\}} in \texttt{\{context\}}.

\textbf{Candidate Pool:} \texttt{\{list\_of\_genes\}}

\textbf{Hierarchical Search Logic:}
\begin{enumerate}
    \item \textbf{Upstream:} Identify TFs from the pool that directly regulate \texttt{\{gene\}}.
    \item \textbf{Downstream:} Identify primary targets of \texttt{\{gene\}} (if it is a TF/signaling node).
    \item \textbf{Complexation:} Identify proteins that form stable, non-redundant complexes with the target.
\end{enumerate}

\textbf{Output Format (JSON):}
\begin{verbatim}
{
  "regulators": [
    {"gene": "SYMB", "confidence": 0-100, "type": "TF/Target/Partner"}
  ],
  "reasoning": "...",
  "mechanism": "Brief description of the circuit."
}
\end{verbatim}
\end{tcolorbox}

%------------------------------------------------------------------------------
\subsection{Task 2: Active Experimental Design}\l
%------------------------------------------------------------------------------
\subsubsection{Step 1: Iterative Master Regulator Selection}
\begin{tcolorbox}[colback=purple!5!white, colframe=purple!75!black, title=Step 1: Iterative Candidate Identification]
\textbf{Context:} Active discovery of the \texttt{\{cell\_line\}} regulatory network.
\textbf{Objective:} Select the next 10 most informative "Master Regulators" to perturb.
\textbf{Iterative Constraints:} 
\begin{enumerate}
    \item \textbf{Novelty:} Prioritize genes not in the already perturbed set: \texttt{\{perturbed\_list\}}.
    \item \textbf{Regulatory Reach:} Select nodes predicted to have maximum transcriptomic impact ($>50k$ targets).
    \item \textbf{Exclusivity:} You MUST select ONLY from the available pool: \texttt{\{gene\_list\_sample\}...}
\end{enumerate}
Return strictly a valid JSON array of 10 gene symbols: \texttt{["REG1", "REG2", ..., "REG10"]}
\end{tcolorbox}
\subsubsection{Step 2: Evidence-Grounded Target Mapping}
\begin{tcolorbox}[colback=orange!5!white, colframe=orange!75!black, title=Mechanistic Verification (Batch)]
Map the primary downstream targets for the candidates: \texttt{\{batch\_list\}}.
\textbf{Requirement:} For each regulator, provide a confidence interval grounded in literature (ChIP-seq, Perturb-seq, RNA-seq). 
\textbf{Confidence Calibration:}
1.0 (Direct cell-type specific evidence); 0.7 (Lineage-wide evidence); 0.4 (Computational/Motif prediction only).
\textbf{Output Format (JSON):}
\begin{verbatim}
{
  "REGULATOR_X": {
    "targets": ["T1", "T2"],
    "confidence": 0.9,
    "logic": "Repression/Activation",
    "evidence_note": "A summary of the supporting literature."
  }
}
\end{verbatim}
\end{tcolorbox}

%------------------------------------------------------------------------------
\section{Consensus Aggregation Protocol}
%------------------------------------------------------------------------------

For the \textsc{MechPert} model, we execute 3 independent reasoning chains with temperature $T=0.7$. The final regulator set is computed as:
\begin{equation}
    w_r = \sum_{i=1}^{3} \mathbb{1}[r \in \mathcal{R}_i] \cdot c_{i,r}
\end{equation}
where $\mathcal{R}_i$ is the regulator set from chain $i$ and $c_{i,r} \in [0,1]$ is the self-reported confidence for regulator $r$ in chain $i$ (normalized from the 0--100 scale). Regulators are ranked by $w_r$ and the top-$k$ are selected.

For \textbf{Binary Consensus}, we set $c_{i,r} = 1$ for all regulators, effectively counting votes:
\begin{equation}
    w_r^{\text{binary}} = \sum_{i=1}^{3} \mathbb{1}[r \in \mathcal{R}_i]
\end{equation}

\end{document}